\documentclass[letterpaper, 10 pt, journal, twoside]{ieeetran}
\usepackage{url}
\usepackage{xcolor}
\usepackage{stfloats}
\usepackage{wrapfig}
\usepackage{multirow}
\usepackage{subcaption}
\usepackage{rotating}
\usepackage{booktabs}
\usepackage{algpseudocode}
\usepackage{booktabs}
\usepackage{float}
\usepackage{graphicx}
\usepackage{amsmath}
\usepackage{amssymb}
\usepackage[demo,abs]{overpic}
\usepackage{xcolor}
\usepackage{caption}
\usepackage{colortbl}
\usepackage{xcolor}  
\let\labelindent\relax 
\usepackage{enumitem}
\usepackage[ruled, lined, linesnumbered, commentsnumbered, longend]{algorithm2e}
\usepackage{hyperref}
\usepackage{soul}
\definecolor{mypink}{rgb}{1, 0., 0.7} 
\setstcolor{mypink}

\hypersetup{
    colorlinks=false,  
    pdfborder={0 0 0}  
}
\definecolor{mydarkgray}{RGB}{46, 46, 46}

\usepackage{colortbl}
\usepackage{multirow}
\usepackage[para]{footmisc}
\usepackage{amsmath}
\usepackage{amssymb}
\usepackage{dsfont}
\usepackage{etoolbox}
\usepackage{color}
\usepackage{ulem}

\newif\ifshowedits

\newcommand{\addeditor}[3]{%
  \definecolor{#1color}{rgb}{#3}
  \expandafter\newcommand\csname #1\endcsname[1]{%
  \ifshowedits
    {\color{#1color} ##1}%
  \else
    {##1}%
  \fi
  }%
  \expandafter\newcommand\csname #1rmk\endcsname[1]{%
  \ifshowedits
    {\color{#1color} {\bf [#2: ##1]}}
  \fi
  }%
  \expandafter\newcommand\csname #1rpl\endcsname[2]{%
  \ifshowedits
    {\color{#1color} ##1 \sout{##2}}
  \else
    {##1}
  \fi
  }%
}

\newcommand{\createtextvar}[1]{
  \expandafter\newcommand\csname #1\endcsname{%
  {\text{#1}}
}%
}
\newcommand{\mycomment}[1]{}

\newcommand{\calD}{{\cal D}}

\newcommand{\calG}{{\cal G}}
\newcommand{\calH}{{\cal H}}
\newcommand{\calI}{{\cal I}}

\newcommand{\calL}{{\cal L}}
\newcommand{\calM}{{\cal M}}

\newcommand{\calO}{{\cal O}}
\newcommand{\calP}{{\cal P}}
\newcommand{\calQ}{{\cal Q}}

\newcommand{\calS}{{\cal S}}

\newcommand{\IR}{{\mathds{R}}}

\newcommand{\vcomment}[1]{}

\usepackage{graphicx}
\usepackage{etoolbox}

\makeatother

\begin{document}
\author{Asma Brazi$^{1,2}$, Boris Meden$^{1}$, Fabrice Mayran de Chamisso$^{1}$, Steve Bourgeois$^{1}$, Vincent Lepetit$^{2}$\\ $^{1}$Université Paris-Saclay, CEA, List,\\ $^{2}$LIGM, Ecole des Ponts, Univ Gustave Eiffel, CNRS, Marne-la-vallee, France}%

\title{Corr2Distrib: Making Ambiguous Correspondences an Ally to Predict Reliable 6D Pose Distributions}
\maketitle
\begin{abstract}
We introduce Corr2Distrib, the first correspondence-based method which estimates a 6D camera pose distribution from an RGB image, explaining the observations. Indeed, symmetries and occlusions introduce visual ambiguities, leading to multiple valid poses. While a few recent methods tackle this problem, they do not rely on local correspondences which, according to the BOP Challenge, are currently the most effective way to estimate a single 6DoF pose solution. Using correspondences to estimate a pose distribution is not straightforward, since ambiguous correspondences induced by visual ambiguities drastically decrease the performance of PnP. With Corr2Distrib, we turn these ambiguities into an advantage to recover all valid poses. Corr2Distrib first learns a symmetry-aware representation for each 3D point on the object's surface, characterized by a descriptor and a local frame. This representation enables the generation of 3DoF rotation hypotheses from single 2D-3D correspondences. Next, we refine these hypotheses into a 6DoF pose distribution using PnP and pose scoring. Our experimental evaluations on complex non-synthetic scenes show that Corr2Distrib outperforms state-of-the-art solutions for both pose distribution estimation and single pose estimation from an RGB image, demonstrating the potential of correspondences-based approaches.
\end{abstract}   
\section{Introduction}
\label{sec:intro}
\begin{figure}[t!]
    \centering
    \begin{overpic}[scale=0.17,unit=1mm]{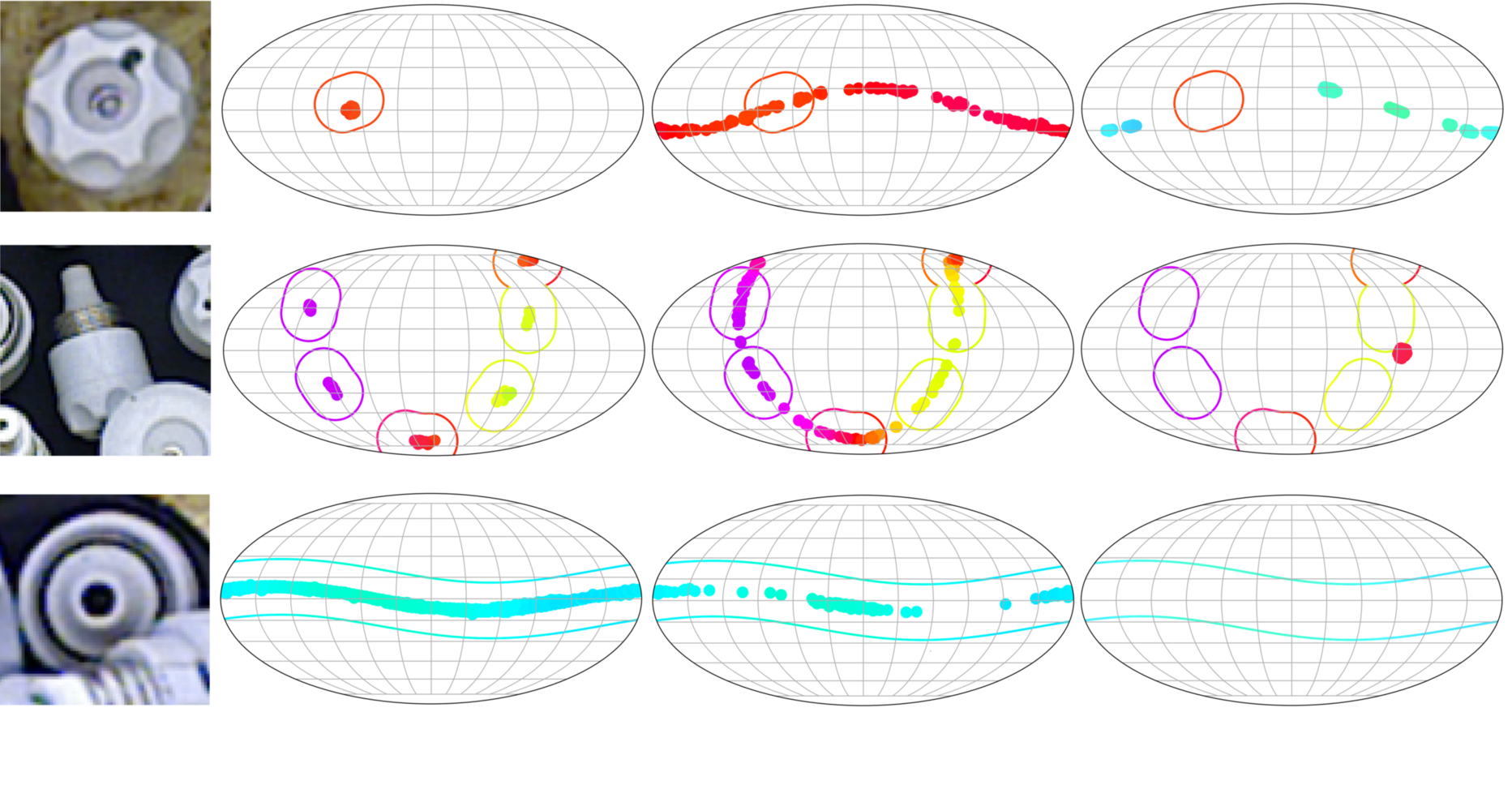}
     \put(1.5, 0.5){\small{Image}}
     \put(13, 0.5){\small{Corr2Distrib (ours)}}
     \put(40, 0.5){\small{LiePose~\cite{hsiao2024confronting}}}
     \put(62, 0.5){\small{SpyroPose~\cite{haugaard2023spyropose}}}
    \end{overpic}
\caption{\textbf{Pose distribution comparison of our method with SpyroPose~\cite{haugaard2023spyropose} and LiePose~\cite{hsiao2024confronting} on the T-LESS dataset~\cite{hodan2017t}}. We illustrate three cases of the object 1 from top to bottom: no symmetry, six-fold rotational symmetry, and continuous rotational symmetry. Rotations are shown in 2D (Mollweide projection), with color indicating tilt direction and marker size representing pose probability ~\cite{murphy2021implicit}. The circles represent ground truth rotations.Our correspondence-based method achieves state-of-the-art results compared to direct methods (SpyroPose and LiePose).}
\label{fig:teaser_fig}
\end{figure}
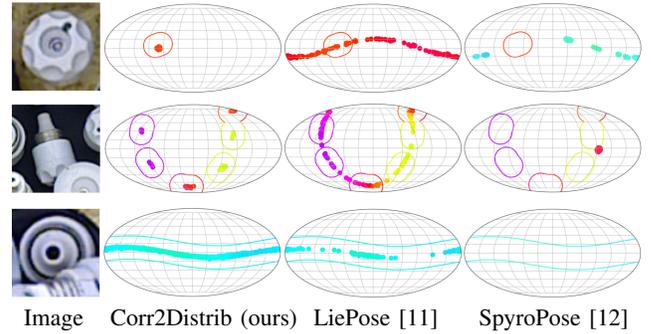

\IEEEPARstart{I}{n} both robotics and augmented reality, estimating the six degrees-of-freedom (6DoF) object pose, meaning its 3D position and 3D orientation, is a crucial task. It enables the transfer of prior knowledge related to the object, such as grasping points, onto the observed scene. It is a key component of various downstream applications, including grasping \cite{wang2019densefusion,xiang2017posecnn,naik2024robotic}, assembly \cite{agin1980computer,zhang2011vision}, bin picking \cite{oh2012stereo,jiang2020depth}, quality control \cite{brito2020machine,andersen2019self}.
Depending on the object's nature (e.g., symmetries) and observation conditions (e.g., occlusions, distance), its observation from sensors can exhibit similar appearances from different viewpoints \cite{manhardt2019explaining,hsiao2024confronting}, yielding multiple or even an infinite number of valid poses as illustrated in Figure \ref{fig:teaser_fig}.  In such cases, knowing whether or not a multiplicity of solutions exists can be extremely valuable for the final application. For example, in robotic grasping, some valid poses can be reachable while others are not, depending on the robot’s kinematic and the surrounding environment. Similarly, in assembly control, pose ambiguity (e.g., due to hidden disambiguating elements) prevents undue validation or invalidation of the assembly. 

However, estimating the full set of possible poses presents a significant challenge, stemming from several critical factors. Unlike most existing works \cite{haugaard2022surfemb,labbe2022megapose,sundermeyer2020augmented,su2022zebrapose,zakharov2019dpod,wang2021gdr,park2019pix2pose,maji2024yolo}, which output a single pose even though multiple poses are valid, estimating a pose distribution requires a fine understanding of the object geometry and its visible parts to decide whether to output a unique, multiple, or infinite solutions.

In this paper, we attempt to solve the problem of estimating the object's 6D pose distribution from a single RGB image. To our knowledge, only two methods explicitly address this problem: a viewpoint classification approach ~\cite{haugaard2023spyropose} and a regression approach~\cite{hsiao2024confronting}. Motivated by the success of correspondence-based single-pose methods like GPose~\cite{zhang2023gpose} in the BOP Challenge \footnote{\url{https://bop.felk.cvut.cz/leaderboards/}}~\cite{hodavn2020bop}, we introduce the first 6D pose distribution method based on 2D-3D correspondences.

Our solution involves learning a symmetry-aware 3D representation, consisting of a field of local descriptors and local coordinate frames respecting the object symmetries.
This representation is learnt simultaneously from RGB images and the object's 3D model. This allows to estimate a 3DoF rotation from a single correspondence, transforming the ambiguous correspondences inherent in symmetrical objects into an advantage to recover the pose distribution, whereas it induces a combinatorial explosion in other correspondence-based solutions\footnote{Single pose estimators typically combine the Perspective-3-Point (P3P) approach \cite{gao2003complete} with a RANSAC-like algorithm \cite{fischler1981random} to handle the probability $p$ to sample a correct correspondence. For non-ambiguous objects,  the probability of sampling a correct triplet is $p^3$ since inlier correspondences are necessarily consistent with each other. However, it decreases to  $p^3/n^2$ when the object has an apparent $n-$fold symmetry since inlier correspondences are not necessarily consistent with each other.}.
We evaluate our method using the newly introduced BOP-Distrib \cite{bopd2024} annotations and protocol, the first and only non-synthetic benchmark that provides pose distribution ground truth and metrics that account for scene conditions and object symmetries. Unlike the BOP~\cite{hodan2024bop} ground truth pose distribution, which are defined independently of the scene configuration, BOP-Distrib dynamically adjusts it based on environmental conditions. Our method achieves state-of-the-art results, demonstrating the effectiveness of correspondence-based approaches for this task.

In summary, our contributions are:
\begin{itemize}[left=0pt]
    \item A learning process producing a symmetry-aware object representation that can be inferred from RGB images.
    \item The first correspondence-based method estimating a 6D pose distribution of an object from a single RGB image while keeping combinatorics tractable.
    \item Thorough evaluations of the method, displaying strong results in comparison to state-of-the-art approaches.
\end{itemize}
\section{Related Work}
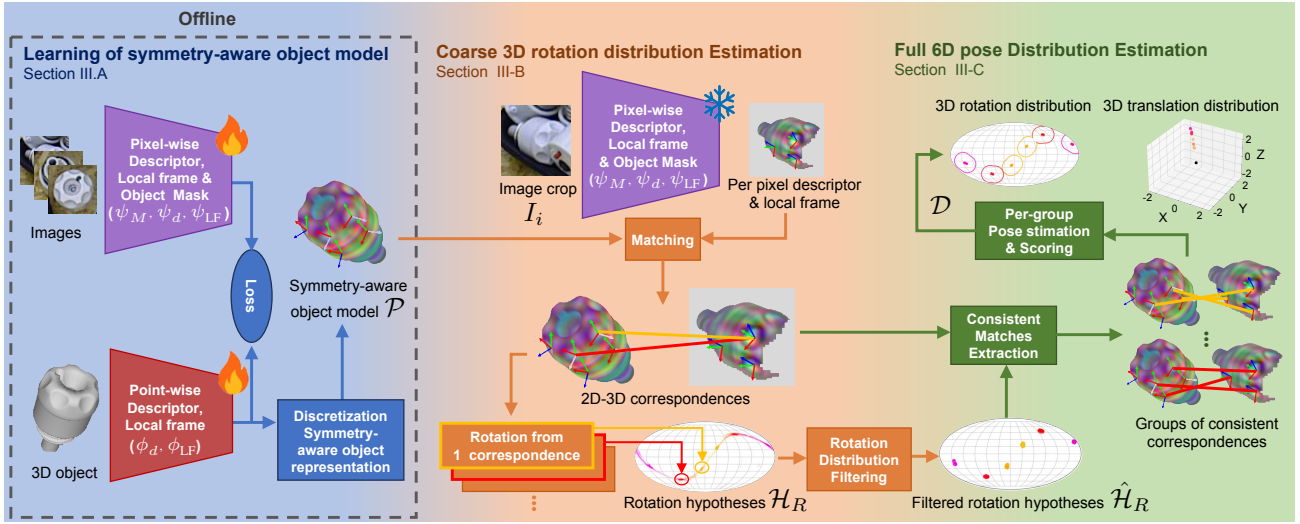
\begin{figure*}[ht!]
\vspace{0.3cm}
    \centering
    \begin{overpic}[scale=0.49,unit=1mm]{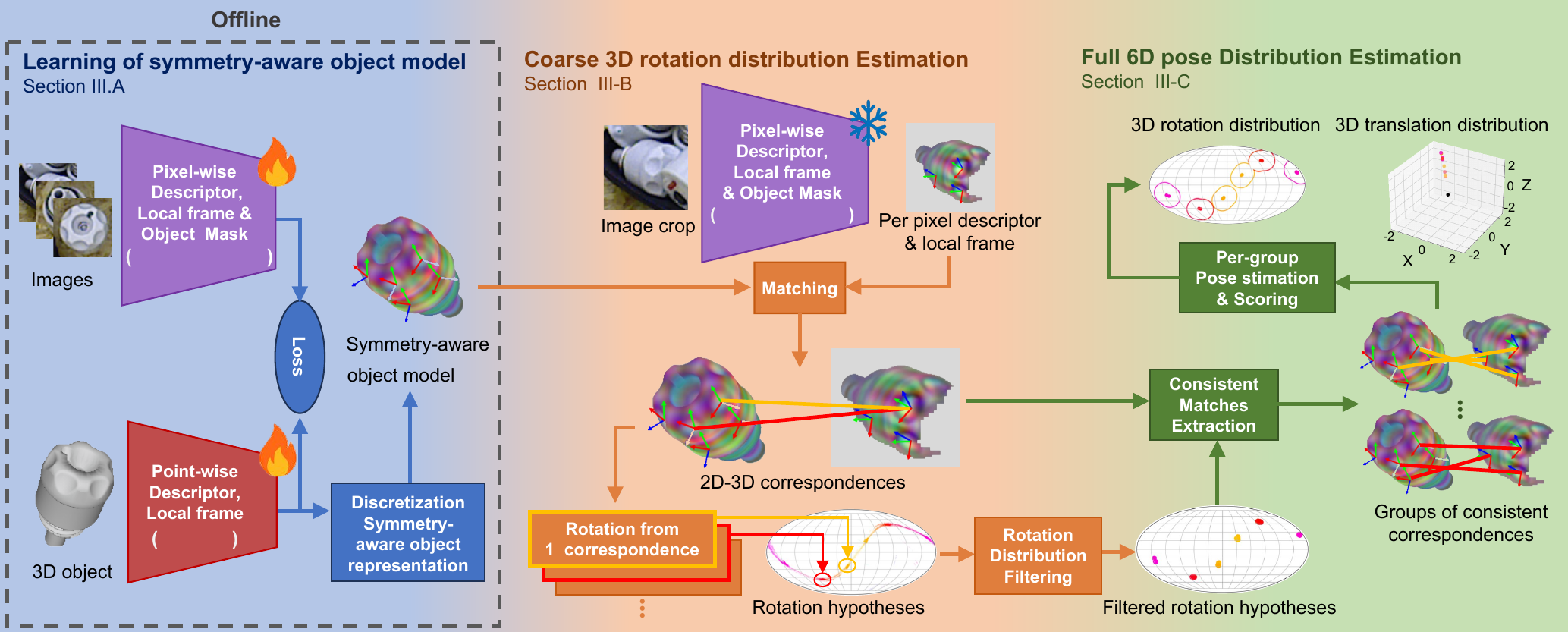}
     \put(14.6, 40.5){\color{white}\scriptsize{$\psi$}\tiny{$_M$}, \scriptsize{$\psi$}\tiny{$_d$}, \scriptsize{$\psi$}\tiny{$_{\text{LF}}$}}
     \put(17.4, 9.5){\color{white}\scriptsize{$\phi$}\tiny{$_d$}, \scriptsize{$\phi$}\tiny{$_{\text{LF}}$}}
     \put(78.3, 45){\color{white}\scriptsize{$\psi$}\tiny{$_M$}, \scriptsize{$\psi$}\tiny{$_d$}, \scriptsize{$\psi$}\tiny{$_{\text{LF}}$}}
     \put(50.5, 27.4){$\calP$}
     \put(69, 40){$I_i$}
     \put(101.5, 2){$\calH_R$}
     \put(147, 2){$\hat{\mathcal{H}}_R$}
     \put(123, 41){$\mathcal{D}$}
    \end{overpic}
    \caption{\textbf{Overview of Corr2Distrib.} We first learn offline a  Discrete Symmetry-Aware object model (Section \ref{sec:learn_symmetry_aware_rep}), consisting in a descriptor and a local frame per object surface point. At inference, given an image crop, we estimate for each pixel a descriptor and a local frame orientation relative to the camera. Then, we establish many-to-many 2D-3D correspondences between image pixels and symmetry-aware model points. From each correspondence, we estimate a 3D rotation, resulting in a set of rotation hypotheses (Section~\ref{sec:equi_pose_hypothesis}). These rotations are filtered and leveraged to group geometrically consistent correspondences. Finally, we estimate a 6D pose from each group of correspondences and score it to generate the final object pose distribution (Section \ref{sec:poseDistribEstim}). Ground truth rotations appear as colored circles, and translation as a black star.}
\label{fig:principal_pipeline}
\end{figure*}
\label{sec:related_work}
\paragraph*{\textbf{Single 6D pose estimation}}
While 6D pose estimation can yield one, many, or even an infinite number of solutions, single-pose estimation methods only output one solution.
\\
To achieve this, indirect approaches, also called correspondence-based approaches, typically follow a two-stage process: they first establish 2D-3D correspondences between 2D image pixels and 3D object points; then, they use these correspondences to compute the 6D pose, often via the PnP algorithm or learning-based methods. Correspondences can be established through:
\begin{itemize}[left=0pt]
\item Given a set of predefined 3D keypoints in a 3D object model, learning to predict their 2D locations in the image using per-keypoint heatmap estimation \cite{rad2017bb8}, keypoint coordinate regression \cite{maji2024yolo}, or a diffusion process \cite{xu20246d-diff}. 
\item Learning to predict the 3D coordinates of each 2D pixel in the object's coordinate system, either by directly regressing the Normalized Object Coordinate System (NOCS) \cite{park2019pix2pose,wang2021gdr}, indirectly through the UV coordinates of a 3D point \cite{zakharov2019dpod} or via any bijective mapping between identifiers and object surface points \cite{su2022zebrapose}.
\item Learning to match image pixels and 3D object points by assigning identical descriptors to corresponding object points and pixels \cite{florence2018dense,haugaard2022surfemb}. Unlike methods that predict 3D coordinates, this approach effectively handles multiple correspondences for visually ambiguous object points by assigning the same descriptor to symmetry-invariant ones.
\end{itemize}

On the other hand, direct approaches estimate the 6D pose directly from the entire image with several variants:
 \begin{itemize}[left=0pt]
    \item Directly regressing 6D pose parameters from the image \cite{kendall2015posenet,xiang2017posecnn,tekin2018real,labbe2020cosypose,maji2024yolo}.
     \item Approaching the problem as a viewpoint classification problem, discretizing the pose space into a set of viewpoints and classifying the current image accordingly. This can be done by inferring the probability distribution among the viewpoints directly from the input image~\cite{kehl2017ssd} or by comparing the image with templates associated to the viewpoints~\cite{sundermeyer2020augmented,shugurov2022osop,nguyen2022templates,labbe2022megapose}. To compensate for the inaccuracy induced by the pose space discretization, a second stage of pose refinement is usually applied.
 \end{itemize}

\paragraph*{\textbf{6D pose distribution estimation}}
Unlike single 6D pose estimation, 6D pose distribution estimation methods aim to output all the valid poses that may explain the observed image. The resulting pose distribution is shaped by the object's geometry (global symmetries), the viewpoint (self-occlusion-induced symmetries), and the object’s visibility (occlusion-induced symmetries).

Adapting direct approaches based on viewpoint classification to 6D pose distribution is straightforward since methods such as \cite{sundermeyer2020augmented} already provide multiple solutions through a multi-modal probability distribution over $\mathrm{SO}(3))$. However, to improve the precision, recent approaches tried to avoid pose space discretization. To this end, classification approaches introduced the concept of an implicit function that, for any given orientation \cite{murphy2021implicit,hofer2023hyperposepdf,cai2022sc6d}, directly maps an encoding of the entire object image onto the probability associated with this orientation. Similarly, template-based approaches replaced explicit templates generation by learning an implicit function that, for any pose, maps image features corresponding to the 2D projection of a specific set of 3D object points (the pattern) to a template matching score \cite{iversen2022ki,haugaard2023spyropose}. 

Alternatively, a direct approach regresses the set of plausible poses. \cite{manhardt2019explaining} regresses a fixed number of pose hypotheses, which may not fully capture the entire range of plausible poses, particularly in scenarios involving continuous symmetry. Recently, \cite{hsiao2024confronting} proposed the first diffusion-based method for $\mathrm{SE}(3)$ to implicitly represent the pose distribution. To recover the set of all possible poses, it requires sampling multiple times from different noisy starting points.

While all SOTA methods are direct approaches, our solution is the first correspondence-based method that exploits ambiguous 2D-3D correspondences to estimate a 6D pose distribution. Unlike \cite{haugaard2023spyropose}, our solution does not require sampling the pose space. In addition, it outputs an explicit set of poses with associated confidence scores, unlike \cite{hsiao2024confronting}.
\section{Method}
\label{sec:method}
\newcommand{\hR}{{\hat{R}}}
\newcommand{\hX}{{\hat{X}}}
\newcommand{\xX}{{x \leftrightarrow X}}

As illustrated in Figure \ref{fig:principal_pipeline}, our method Corr2Distrib relies on three stages to estimate the object pose distribution $\calD$ given an RGB image crop and a 3D object model.

The first stage, achieved offline, consists of learning a symmetry-aware representation of the 3D object whose projection from a given viewpoint can be inferred from the corresponding RGB image using a neural network (Section \ref{sec:learn_symmetry_aware_rep}). 
We define symmetry-awareness as the ability of an object representation to capture, for any two object points, whether there is an object symmetry (intrinsic or induced) aligning one point to the other, as well as the rigid transformation that achieves this alignment.

The second stage, achieved online, leverages the symmetry-aware representation to coarsely estimates the rotational component of the pose distribution $\mathcal{D}$  from an RGB image. The pixels of the image are matched to the symmetry-aware representation, generating a set of 2D-3D correspondences, each associated with a 3D rotation (Section \ref{subsection:local_frames}). For symmetrical objects, a pixel may correspond to multiple 3D points. While this ambiguity is challenging for classical PnP methods, it provides a set of rotations that yield a coarse approximation of the rotation distribution.

The final stage extends and refines the coarse rotation distribution into a fine 6D pose distribution (Section \ref{sec:poseDistribEstim}).
 
\subsection{Symmetry-aware object representation learning}
\label{sec:learn_symmetry_aware_rep}
In our symmetry-aware representation, each object point is assigned a descriptor and a local frame, ensuring that points related by symmetry share the same descriptor and have local frames aligned through the corresponding symmetry transformation.

For asymmetrical objects, this assignment is straightforward~\footnote{The 3D point position can be used as an identifier, while the same local frame can be assigned to any point}. However, for objects with intrinsic symmetries, the assignment becomes significantly more complex, especially when visual symmetries are influenced by partial occlusions or observation distance, for example.

To address this problem, we propose to learn both the descriptors and local frames assignment from a collection $\calI=\{I_i\}$ of localized cropped images of the object in various poses. The assignment is learnt implicitly by training a neural network to predict, from any given image of the object, the descriptor and the local coordinate frame of the 3D points underlying the pixels. On one side, object symmetries ---whether intrinsic or induced--- imply that the image of the object remains invariant under certain rigid 3D transformations. These symmetries introduce visual ambiguities among images in $\calI$, which, by design, force the neural network to predict the same descriptor and local frame for different 3D points, even when observed from different viewpoints. This property behaves as an implicit supervision signal to recover symmetries among the object points. 
On the other side, a 3D point of the object is observable in multiple images within $\calI$. The poses associated with these images can be used to supervise the multi-view consistency of the descriptor and local frame inferred for this 3D point.

\subsubsection{Descriptors learning}
\label{subsubsec:descriptorlearning}
We consider two functions: 
\begin{itemize}[left=0pt]
    \item $\phi_d: \IR^3 \rightarrow E_{\text{desc}}$  that assigns to any 3D point $X$ a descriptor $\phi_d(X)\in E_{\text{desc}}$, with $E_{\text{desc}}$ corresponding to the descriptor space.
    \item $\psi_d: \mathbb{R}^{3 \times W \times H} \rightarrow E_{\text{desc}}^{W \times H }$ that maps a cropped RGB image $I$ of an object to a descriptor image, where each pixel $x$ is associated with a descriptor denoted $\psi_d(I,x)$for simplicity.
\end{itemize}
To ensure proper descriptor assignment, we need:
\begin{itemize}[left=0pt]
    \item Multi-view consistency: $\phi_d(X)$ and $\psi_d(I,x)$ should be similar when $X$ is the underlying 3D object point of the pixel $x$ in image crop $I$. 
    \item Descriptors as identifiers: The similarity between $\phi_d(Y)$ and $\psi_d(I,x)$ should be minimal when $X \neq Y$, with $X$ the underlying 3D object point of the pixel $x$ in the image crop $I$. 
\end{itemize}

To reach this objective, we train  $\psi_d$ and $\phi_d$ simultaneously using the contrastive loss $\calL_{\text{desc}}$ based on InfoNCE \cite{haugaard2022surfemb}:
\begin{equation}
\label{eq:desc}
    \calL_{\text{desc}} = -\frac{1}{\mid\calM\mid}  \sum_{x \in \calM} \text{sim}_{\text{desc}}(\psi_d(I,x),h(x)),
\end{equation}    
where $\text{sim}_{\text{desc}}$ measures the similarity between the descriptors of a pixel $x$ and a point $X$:
\begin{equation}
\hspace{-0.3cm} 
    \text{sim}_{\text{desc}}(I,x,X) = \log\frac{
        \exp(\psi_d(I,x)^T.\phi_d(X)))
    }{
        \sum_{Y \in \calO} \exp(\psi_d(I,x)^T.\phi_d(Y)))
    };
\end{equation}

where $\calM_i$ is the mask of the object in image $I$, $\calO$ the 3D surface points of the object, $h(x) \in \calO$ the 3D object point underlying the pixel $x$ in image $I$, and  $\text{sim}_{\text{desc}}$ the function measuring the similarity between the descriptor associated to the pixel $x$ of $I$ and the one associated with the 3D point $X$ of $\calO$.

Although this loss does not explicitly handle object symmetries, these symmetries will naturally be respected due to the visual ambiguities they introduce. For symmetrical objects, a 3D point may correspond to multiple pixels in different orientations and vice versa. Since symmetrical poses cannot be distinguished from the image alone, image descriptors are naturally constrained to take the same values for symmetric points or those appearing symmetric due to occlusions. This makes the descriptor field  symmetry-aware. This constraint is transferred from the image domain to the object's representation through the InfoNCE loss. This happens despite the loss penalizing different model points with similar descriptors as non-matches, thanks to the imbalanced handling of matches and non-matches in Equation \ref{eq:desc}.

Additionally, we train a function $\psi_M$ on the images $\calI$ that returns the object mask $\calM$ from an RGB image $I$.

\subsubsection{Symmetry-aware local frame learning}
\label{subsection:local_frames}
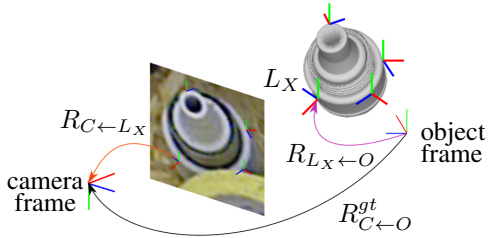
\begin{figure}[!bt]
\vspace{0.25cm}
     \centering
     \begin{overpic}[scale=0.26,unit=1mm]{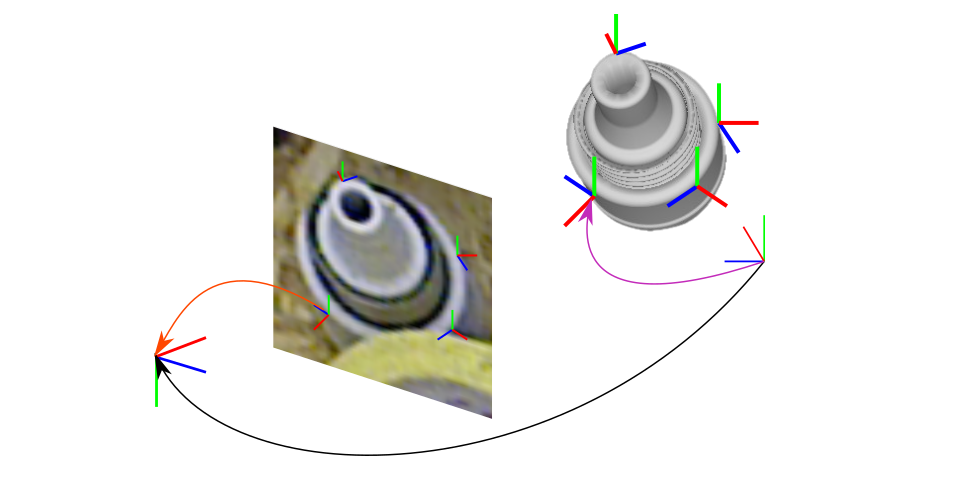}
     \put(0, 8.5){camera}
     \put(1, 5.5){frame}
     \put(55, 15){object}
     \put(55, 12){frame}
     \put(37, 12){$R_{L_X \leftarrow O}$}
     \put(7, 16.5){$R_{C \leftarrow L_X}$}
     \put(44, 4){$R^{gt}_{C \leftarrow O}$}
     \put(34, 22){$L_X$}
    \end{overpic}
     \caption{\textbf{Visualization of the transformations used in local frames learning.} The different frames are: camera frame, object frame, and local frames (L$_X$). For a given 2D-3D correspondence $(x,X)$, $\psi_{LF}$ returns \(R_{C \leftarrow L_X}\) and $\phi_{LF}$ returns \(R_{L_X \leftarrow O}\). The combination of these rotations provides the desired rotation \(R_{C \leftarrow O}\), which is supervised by \(R^{gt}_{C \leftarrow O}\).}
     \label{fig:method_frames}
\end{figure}

We consider two functions: 
\begin{itemize}[left=0pt]
    \item $\phi_{LF} : \mathbb{R}^3 \rightarrow \mathrm{SO}(3)$ that assigns, to any object surface point $X \in O$, the rotation $R_{L_X \leftarrow O}$ of the local  frame $L_X$ assigned to that point $X$ in the object coordinate frame.
    \item $\psi_{LF}$ : $\mathbb{R}^{3 \times W \times H } \rightarrow \mathrm{SO}(3)^{W \times H }$ that, given a cropped RGB image $I$, infers for each pixel $x\in I$ the relative rotation $\psi_{LF}(I,x)$ = $R_{C \leftarrow L_X}\in \mathrm{SO}(3)$ between the local coordinate frame $L_X$ of the point $X$ of the 3D object underlying $x$ and the camera $C$ that captured image $I$.
\end{itemize}
The function $\psi_{LF}$ should respect the multi-view consistency: given a  3D object point  $X\in\calO$,  in any image $I$ observing $X$ at pixel coordinate $x$, the relationship $R_{C \leftarrow L_X}.R_{L_X \leftarrow O} =R^{gt}_{C \leftarrow O}$ holds, with $R^{gt}_{C \leftarrow O}$ corresponding to the ground truth orientation of the camera $C$ associated to $I$ as illustrated in \autoref{fig:method_frames}. 
To this end, the functions $\psi_{LF}$ and $\phi_{LF}$ are trained jointly with the following $\calL_{LF}$ loss:  
\begin{equation}
\calL_{LF} = \frac{1}{\mid\calM\mid} \sum_{x\in \calM}d_{\text{ang}}(R^{gt}_{C\leftarrow O},\psi_{LF}(I,x).\phi_{LF}(X)),
\label{eq:lossrot}
\end{equation}
with $\calM_i$ the object mask in image $C$,  $d_{\text{ang}}$ the distance between two rotations, expressed as the relative rotation angle, $X\in \calO$ the 3D object point that underlies the pixel $x$ of image $I$.

Similarly to descriptors learning, $\calL_{LF}$ operates without explicit guidance regarding object symmetries, yet these symmetries naturally arise from the visual ambiguities they induce. Additionally, local frames learnt from the image are necessarily symmetry-invariant. When they are matched to local frames learnt from the object model, $\calL_{LF}$ constrains local frames of equivalent object points under a symmetry to be symmetry-equivariant thanks to the match/non-match imbalance once again.

Since we train on cropped images, objects with similar allocentric orientation typically exhibit similar appearance,  regardless of the object's location in the image and the associated distortion caused by parallax \cite{kundu20183d}. Hence, using per-pixel allocentric reference frame is more suitable to learn rotations. For performance reasons, during training, the ground truth rotation is converted to allocentric coordinates and at inference, the predicted rotation is converted to egocentric coordinates relative to the camera. 

\subsubsection{Discretized Symmetry-aware object representation}
\label{subsubsec:dsamodel}
Once the local frame and  descriptor fields are learnt, the resulting implicit representation of the object (i.e., $\phi_d$ and $\phi_{LF}$) is converted into an explicit representation that can be more easily and efficiently exploited in the following. 

A 3D point cloud is sampled evenly from the object's surface, with each point associated with its  corresponding descriptor and local coordinate frame in the implicit fields. The resulting explicit and symmetry-aware object model is denoted as $\calP$. For simplicity, given a 3D point $X$ in $\calP$, we refer to its associated descriptor (resp. local frame) as  $\calP_d(X)$ (resp. $\calP_f(X)$). 

\subsection{Coarse 3D rotation distribution from ambiguous 2D-3D correspondences}
\label{sec:equi_pose_hypothesis}
The aim of this first step of the online process is to establish 2D-3D correspondences between the image crop $I$ and the 3D symmetry-aware object model $\calP$, while also providing an initial estimate of the rotational part of the 6D pose distribution.
\subsubsection{Sampling rotation hypotheses}
We aim to estimate an initial estimate of the 3D rotation distribution. As depicted in  \autoref{fig:principal_pipeline}, given an image crop $I$, we first compute the object mask, the image descriptors, and the local frames expressed in the camera frame using  $\psi_M$, $\psi_d$ and $\psi_{LF}$ respectively. For each pixel $x$ in $\psi_M(I)$, we compare its descriptor $\psi_d(I,x)$ with all the descriptors of $\calP$. Rather than selecting only the nearest 3D point $\hX$ in descriptor space as usually done, we leverage the symmetry-awareness of $\calP$ to associate $x$ with  $\calS(I,x)$, the set of 3D points of $\calP$  that can underlie $x$ considering the object symmetries:
\begin{equation}
\calS(I,x)=\{X \in\calP | \text{sim}_{\text{desc}}(I,x,X)<\delta\},
\end{equation}
with $\delta=\tau_{desc}*\text{sim}_{\text{desc}}(I,x,\hX)$ a threshold computed automatically as a fraction $\tau_{desc}$ of the maximal similarity.

Then, we leverage the ability to estimate the camera rotation relative to the object coordinate frame for each 2D-3D correspondence between a pixel $x$ and a 3D point from $\calS(I,x)$. This allows to derive the set $\calH^x_R$ of 3D rotation hypotheses associated with $x$ as follow:
\begin{equation}
    \calH^x_R=\{ R^X_{C\leftarrow \calO}= R_{C\leftarrow L_X}.R_{L_X \leftarrow \calO}, X\in\calS(I,x)\},
\end{equation}
with  $R_{C\leftarrow L_X}=\psi_{LF}(I,x)$,  and $R_{L_X \leftarrow \calO}=\calP_f(X)$.
The complete set of camera rotation hypotheses, $\calH_R$, for the image crop $I$ is defined as the aggregation of the rotation hypotheses corresponding to each pixel within the object mask: $\calH_R(I)=\{ \calH^x_R, x\in \calM\}$.
\begin{figure}[h!]
    \centering
    \begin{overpic}[scale=0.27,unit=1mm]{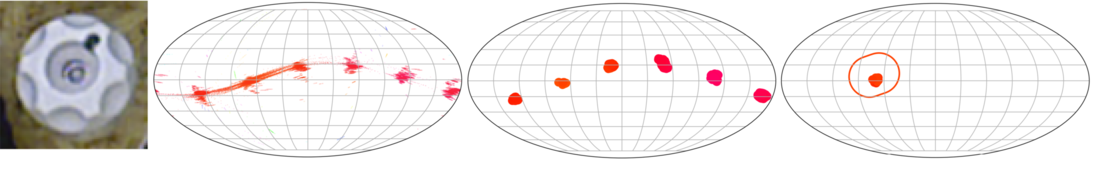}
        \put(1, 0){\footnotesize{Image}}
        \put(12, 0){\footnotesize{Init. rotations $\calH_R$}}
        \put(35, 0){\footnotesize{After filtering $\hat{\calH}_R$}}
        \put(59, 0){\footnotesize{Final rotations}}
    \end{overpic}
      \caption{\textbf{Evolution of the rotation distribution from coarse to fine.} We start by estimating an initial  set of rotation hypotheses from single 2D-3D correspondences (Section \ref{sec:equi_pose_hypothesis}). This set is then  filtered by discarding rotations according to observation conditions and $\mathrm{SO}(3)$ density regions (Section \ref{sec:rotation_refinement}). Finally, we refine the distribution using PnP-RANSAC and pose scoring (section \ref{sec:poseDistribEstim}).}
      \label{fig:corse_to_fine}
\end{figure}
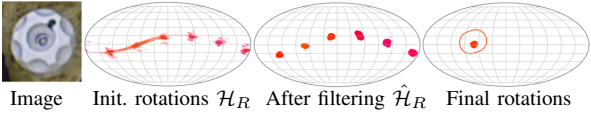
\subsubsection{From rotation hypotheses to rotation distribution}
\label{sec:rotation_refinement}
The set of rotation hypotheses $\calH_R(I)$ is computed by deriving rotations from each pixel independently. This local nature of each hypothesis implies both a non-negligible noise in the estimation and an overestimation of the potential object symmetries, as shown in Figure \ref{fig:corse_to_fine}, since the latter are considered without  accounting for the actual observation conditions  (distance of observation and occlusions) . As illustrated in Figure \ref{fig:principal_pipeline}, the resulting set of rotation hypotheses $\calH_R(I)$ is noisy and includes invalid hypotheses. However, by construction, the set of hypotheses depicts a higher density of hypothesis in the area of $\mathrm{SO}(3)$ corresponding to the valid rotations. Indeed, those regions are the only ones that agglomerate both the rotation hypotheses provided by pixels underlying symmetric object parts but also ones from less ambiguous parts and even unambiguous elements.  

Based on this observation, we propose to filter the rotation hypotheses based on a density criterion. To this end, we discretize the $\mathrm{SO}(3)$ space following the equi-volumetric partition $\calG_k$ introduced in HealPix~\cite{yershova2010generating}, with $k$ the subdivision level of the grid. The discrete density function $\calQ(\calH_R(I),\calG_k,i)$, representing the density of hypotheses of $\calH_R(I)$ in the $i^{th}$ bin of $\calG_k$, is then computed. We finally compute the pruned set of rotation hypotheses as follows: 

\begin{equation}
\hspace{-0.25cm} 
 \hat{\calH}_R(I) = \{ R \in \calH_R(I) \mid  \calQ(\calH_R(I),\calG_k,\zeta(R)) > \tau_{\text{dens}} \},
\end{equation}  
with $\zeta(R)$ the function that returns the bin index of $\calG_k$ containing the rotation $R$, and $\tau_{\text{dens}}$ a density threshold.

\subsection{Full 6D Pose distribution estimation}
\label{sec:poseDistribEstim}
As discussed in the introduction, estimating the full 6D pose distribution directly via PnP-RANSAC across all 2D-3D correspondences implies an expensive combinatorics cost due to the low probability of sampling three mutually consistent inliers. However, while identifying correspondences consistent with a pose is combinatorically expensive, verifying whether a correspondence is consistent with a given rotation is not. Since each rotation in $\hat{\calH}_R(I)$ is associated with a 2D-3D correspondence, we simply group the correspondences whose associated rotation belong to the same bin in $\calG_k$. For each group, we apply PnP-RANSAC and score the resulting pose hypotheses $P_{C\leftarrow\calO}$. Following \cite{haugaard2022surfemb}, we define the scoring in Equation \ref{eq:scoring} as a combination of descriptor similarity $\gamma_{desc}(I, P_{C \leftarrow \mathcal{O}})$ and mask agreement $\gamma_{mask}(I, P_{C \leftarrow \mathcal{O}})$:
\begin{equation}
\gamma(C, P_{C \leftarrow \mathcal{O}}) = 
\gamma_{\text{desc}}(I, P_{C \leftarrow \mathcal{O}}) + \gamma_{\text{mask}}(I, P_{C \leftarrow \mathcal{O}}).
\label{eq:scoring}
\end{equation}
The term $\gamma_{desc}(I, P_{C \leftarrow \mathcal{O}})$ measures the average similarity in descriptor space between each 3D point $X$ and its 2D projection $\pi_{P_{C\leftarrow\calO}}(X)$ under the pose $P_{C\leftarrow\calO}$:
\begin{equation}
\hspace{-0.25cm} 
\gamma_{desc}(I, P_{C \leftarrow \mathcal{O}}) = \frac{1}{\mid\hat{\calP}\mid}\sum_{X\in\hat{\calP}} \text{sim}_{\text{desc}}( I,\pi_{P_{C\leftarrow\calO}}(X),X),
\end{equation}
with $\hat{\calP}$ the visible 3D points of $\calP$ in pose $P_{C\leftarrow\calO}$, $\pi$ the camera projection function.

Even if the descriptors of a point and its 2D projection are similar, there is no guarantee that the 2D projection aligns with the mask. We define the mask score to penalize misaligned points as follows:
\begin{equation}
\gamma_{mask}(I, P_{C \leftarrow \mathcal{O}}) =
\frac{1}{\mid\hat{\calP}\mid}\sum_{X\in\hat{\calP}}\psi_M(I,\pi(P_{C\leftarrow\calO}.X)).
\end{equation}
Notice that measuring the reprojection error in descriptor space improves the robustness to symmetries compared to the classical geometrical reprojection error. 

Finally, we discard any pose with a score lower than the threshold $\tau_{score}\times scoreMax$, with $scoreMax$ the maximal score among the pose hypotheses.
\section{Experiments}
\label{sec:experiments}
\subsection{Implementation details}
Following \cite{haugaard2022surfemb}, we employ a multi-decoder UNet with ResNet18 \cite{he2016deep} as backbone pretrained on ImageNet \cite{deng2009imagenet} for both $\psi_d$ and $\psi_{LF}$, trained jointly on all objects. For $\phi_d$ and $\phi_{LF}$, we adopt an implicit representation based on the SIREN architecture~\cite{sitzmann2020implicit}, with a separate network trained per object. We train $\psi_d$ and $\phi_d$ over 10 epochs, whereas $\psi_{LF}$ and $\phi_{LF}$ are trained over 50 epochs, both on the T-LESS dataset, using synthetic PBR images. We sample up to 50,000 points per object model.

The predicted rotations are expressed using a three-parameter angle-axis representation where the direction of the vector defines the rotation axis, and its magnitude encodes the rotation angle. We choose the angle-axis representation for its compactness to avoid overparameterization, and for its continuity to facilitate the optimization process \cite{zhou2019continuity}. 
Unless specified, hyperparameters are set as follows:  the descriptor threshold $\tau_{desc}$ is automatically adjusted with a ratio of 0.65, the $\mathrm{SO}(3)$ space is discretized using $\calG_4$, the pose hypothesis density threshold, $\tau_{\text{dens}}$, is set to 10, and the scoring threshold $\tau_{score}$ is set to 0.9.

\subsection{Dataset and evaluation protocol}
We evaluate our method on the BOP-Distrib benchmark \cite{bopd2024}, that extends the T-LESS dataset \cite{hodan2017t} with ground truths and metrics adapted to 6D-pose distributions. To our knowledge, it is the only non-synthetic dataset offering 6D pose distribution ground truths while encompassing non-simplistic geometric shapes. The per-image 6D pose-distributions account for symmetries induced by occlusions and observation distance. The benchmark includes 30 industry-relevant objects, with symmetries and challenging occlusion levels.

We use the pose-distribution metrics introduced in BOP-Distrib, evaluating pose distributions in terms of precision and recall according to two distances: Maximum Projective Distance (MPD) for pixel-based errors and Maximum Surface Distance (MSD) for 3D alignment. The analysis should consider precision and recall as a pair when comparing methods. For consistency, we follow BOP-Distrib’s experimental protocol, using ground-truth image crops as input to ensure performance differences reflect the pose-distribution method, not pre-processing steps like 2D object detection.

\subsection{Pose distribution evaluation}
\label{sec:pose_distribution_evaluation}
This experiment provides a comparative evaluation of our solution with current state-of-the-art methods.\\
\textbf{Evaluated methods.} We compare our method Corr2Distrib against LiePose \cite{hsiao2024confronting} and SpyroPose\cite{haugaard2023spyropose}. To avoid biases related to the training data and ensure a fair comparison, all the methods were trained on the same synthetic dataset of T-LESS \cite{hodavn2020bop}.

\textbf{Quantitative results.}
Table \ref{tab:exp_dist_pose} compares our method with LiePose \cite{hsiao2024confronting} and SpyroPose\cite{haugaard2023spyropose}, using results reported in BOP-Distrib \cite{bopd2024}. On average, our method outperforms the other methods in both MPD and MSD. We can observe that Corr2Distrib demonstrates a slight improvement in MPD precision, while maintaining a strong recall. Despite Spyropose achieving a higher MSD precision, it suffers from a very low recall, which makes our method the best overall when considering the precision and recall pair.

\begin{table}[h!]
\centering
\resizebox{\linewidth}{!}{
    \begin{tabular}{l c c c c c}
        \toprule
        Method  & $\mathbf{P_{MPD}}$ & $\mathbf{R_{MPD}}$ & $\mathbf{P_{MSD}}$ & $\mathbf{R_{MSD}}$ & RANK\\
        \midrule
        LiePose \cite{hsiao2024confronting} & 0.612 & \textbf{0.895} & 0.246 & 0.714 & 2\\
        SpyroPose~\cite{haugaard2023spyropose} & 0.559 & 0.555 & \textbf{0.328} & 0.481  & 3\\
        Corr2Distrib (Ours) & \textbf{0.623} & 0.892 & 0.260 & \textbf{0.720} & 1\\
        \bottomrule
    \end{tabular}
}
\caption{\textbf{Evaluation of pose distribution on the dataset T-LESS~\cite{hodan2017t} using BOP-Distrib \cite{bopd2024} evaluation protocol.} We report Precision and Recall with the MPD and MSD scores in comparison with SpyroPose \cite{haugaard2023spyropose} and LiePose \cite{hsiao2024confronting}.} 
\label{tab:exp_dist_pose}
\end{table}

\textbf{Qualitative results}  We provide a qualitative comparison between our method, SpyroPose \cite{haugaard2023spyropose} and LiePose \cite{hsiao2024confronting} for unimodal and multimodal pose distributions in  Figure \ref{fig:teaser_fig}. 

\begin{figure}[h!]
    \centering
    \begin{overpic}[scale=0.285,unit=1mm]{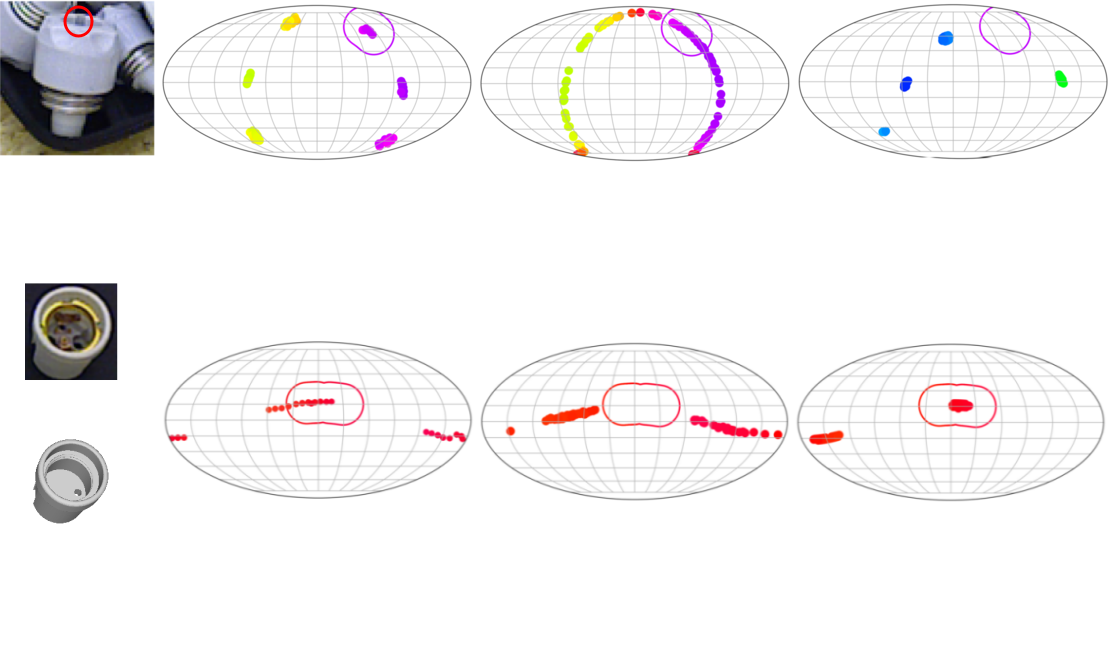}
     \put(1.5, 32.5){\small{Image}}
     \put(1.5, 16.75){\small{Image}}
     \put(1, 5){\small{Object}}
     \put(12, 6){\small{Corr2Distrib (ours)}}
     \put(39, 6){\small{LiePose \cite{hsiao2024confronting}}}
     \put(62, 6){\small{SpyroPose \cite{haugaard2023spyropose}}}
     \put(12, 32.5){\small{Corr2Distrib (ours)}}
     \put(39, 32.5){\small{LiePose \cite{hsiao2024confronting}}}
     \put(62, 32.5){\small{SpyroPose \cite{haugaard2023spyropose}}}
     \put(29, 28){\textbf{\small{Disambiguating element }}}
     \put(27, 1){\textbf{\small{Texture inconsistency impact}}}
    \end{overpic}
    
\caption{\textbf{Failure cases.} The figure illustrates two key sources of error: too small disambiguating element  (first row), and texture inconsistencies between 3D object model and observed  object (second row).}
\label{fig:limitations}
\end{figure}
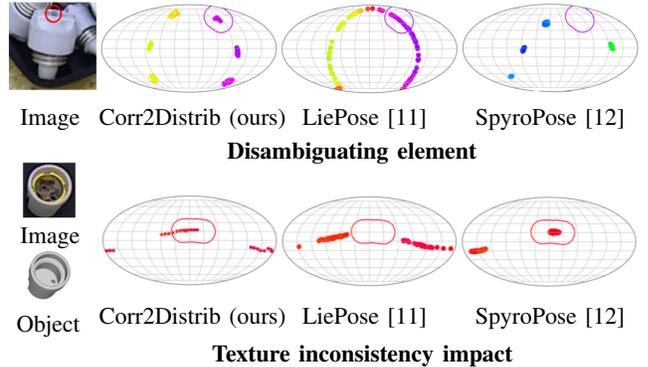

\subsection{Single Pose Evaluation}
\textbf{Evaluated methods.} 
\label{sec:single_pose_evaluation}For a deeper comprehensive accuracy analysis, we evaluate the best pose from the distribution using BOP metrics and BOP-Distrib ground truths, comparing it against the current top-performing single pose  (CosyPose \cite{labbe2020cosypose} and HccePose \cite{yulin2024hcce}) and pose distribution (LiePose \cite{hsiao2024confronting} and SpyroPose \cite{haugaard2023spyropose}) methods. The scores of these baseline methods are taken from BOP-Distrib~\cite{bopd2024}. 

To ensure fair comparison with single pose methods (as pose distribution methods tend to use ground-truth bounding boxes), we also evaluated Corr2Distrib with a state-of-the-art 2D object detector~\cite{zhang2023gdet}.

\textbf{Quantitative results.}
We report the results in Table \ref{tab:exp_single_pose}. We observe  that our method, whether it uses ground truth bounding boxes or inferred ones, outperforms both single pose and distribution methods. We also observe little variation between Corr2Distrib results on both bounding boxes sources.

Considering the negligible nature of the bounding box, we can observe that methods designed to estimate pose distributions significantly outperform all methods designed to estimate a single pose. It suggests that, beyond our method that provides the best performances, pose distribution methods are relevant even to estimate a single pose.

\begin{table}[h!]
\centering
\begin{tabular}{cccccc}
\toprule
Method &Type & BB & MSPD  & MSSD & AVG \\
\hline
HccePose\cite{yulin2024hcce} & Single & \cite{GDRNPPDet}& 0.645 & 0.531 &  0.588   \\
 CosyPose \cite{labbe2020cosypose} & Single & \cite{labbe2020cosypose} &  0.640 & 0.572 &  0.606  \\
 Corr2Distrib (Ours) & Distrib & \cite{zhang2023gdet} & \textbf{0.844} & \textbf{0.591} &  \textbf{0.718}  \\\addlinespace \midrule
SpyroPose \cite{haugaard2023spyropose}& Distrib & GT & 0.736 & \textbf{0.693} &  0.714  \\
 LiePose \cite{hsiao2024confronting} & Distrib &  GT & 0.761 & 0.483 &  0.622  \\
 Corr2Distrib (Ours) & Distrib & GT & \textbf{0.854} &  0.584 & \textbf{0.719} \\
\bottomrule
\end{tabular}
\caption{\textbf{Evaluation of the best pose.}
We compare MSSD and MSPD scores against the top RGB-based single-pose and  distribution methods. Additionally, we specify whether the bounding boxes are ground truth or predicted.
}
\label{tab:exp_single_pose}
\end{table}

\textbf{Run-time.}
We conduct all experiments on a single NVIDIA L40 GPU. The total inference time is 1.408s, with the most time-consuming step being the pose distribution refinement at 1.072s.

\subsection{Ablation Study}
In the following, we derive various experiments to evaluate the sensitivity of our method to its various hyper-parameters. We report the results in Table \ref{tab:ablation}.
\begin{table}[h!]
\centering
\begin{tabular}{c c c c c c}
\toprule
     & & $\mathbf{P_{MPD}}$ & $\mathbf{R_{MPD}}$ & $\mathbf{P_{MSD}}$ & $\mathbf{R_{MSD}}$ \\
    \midrule
    \multirow{4}{*}{\rotatebox[origin=c]{90}{$k$}} &
     3 & \cellcolor{red!35}0.622 & \cellcolor{red!25}0.875 & \cellcolor{red!50}\textbf{0.277} & \cellcolor{red!10}0.625 \\
     &{\color{red}4} & \cellcolor{red!50}\textbf{0.625} & \cellcolor{red!50}\textbf{0.887} & \cellcolor{red!35}0.263 & \cellcolor{red!30}0.714 \\
     &5 & \cellcolor{red!30}0.621 & \cellcolor{red!45}0.886 & \cellcolor{red!25}0.246 & \cellcolor{red!50}\textbf{0.772} \\
     &6 & \cellcolor{red!10}0.608 & \cellcolor{red!10}0.854 & \cellcolor{red!10}0.230 & \cellcolor{red!40}0.755 \\
     
    \midrule
    \multirow{5}{*}{\rotatebox[origin=c]{90}{$\tau_{\text{corr}}$}} &
     0.60 & \cellcolor{red!10}0.621 & \cellcolor{red!50}\textbf{0.892} & \cellcolor{red!10}0.256 & \cellcolor{red!40}0.719 \\
     &{\color{red}0.65} & \cellcolor{red!20}0.623 & \cellcolor{red!50}\textbf{0.892} & \cellcolor{red!20}0.260 & \cellcolor{red!45}\textbf{0.720} \\
     &0.70 & \cellcolor{red!15}0.622 & \cellcolor{red!45}0.891 & \cellcolor{red!20}0.260 & \cellcolor{red!35}0.717 \\
     &0.75 & \cellcolor{red!25}0.624 & \cellcolor{red!40}0.890 & \cellcolor{red!20}0.260 & \cellcolor{red!30}0.714 \\
     &0.80 & \cellcolor{red!30}\textbf{0.626} & \cellcolor{red!35}0.890 & \cellcolor{red!25}\textbf{0.261} & \cellcolor{red!25}0.709 \\

    \midrule
    \multirow{4}{*}{\rotatebox[origin=c]{90}{$\tau_{\text{dens}}$}}
    & 5  & \cellcolor{red!40}\textbf{0.623} & \cellcolor{red!40}\textbf{0.892} & \cellcolor{red!20}0.258 & \cellcolor{red!30}0.719 \\
     &{\color{red}10} & \cellcolor{red!40}\textbf{0.623} & \cellcolor{red!40}\textbf{0.892} & \cellcolor{red!30}\textbf{0.260} & \cellcolor{red!40}\textbf{0.720} \\
     &15 & \cellcolor{red!40}\textbf{0.623} & \cellcolor{red!30}0.891 & \cellcolor{red!25}0.259 & \cellcolor{red!30}0.719 \\
     &20 & \cellcolor{red!40}\textbf{0.623} & \cellcolor{red!30}0.891 & \cellcolor{red!25}0.259 & \cellcolor{red!25}0.718 \\

    \midrule
    \multirow{4}{*}{\rotatebox[origin=c]{90}{$\tau_{\text{score}}$}}
     &0.80  & \cellcolor{red!10}0.572 & \cellcolor{red!50}\textbf{0.912} & \cellcolor{red!10}0.224 & \cellcolor{red!50}\textbf{0.752} \\
     &0.85  & \cellcolor{red!25}0.594 & \cellcolor{red!45}0.904 & \cellcolor{red!20}0.239 & \cellcolor{red!30}0.714 \\
     &{\color{red}0.90}  & \cellcolor{red!35}0.623 & \cellcolor{red!40}0.892 & \cellcolor{red!30}0.260 & \cellcolor{red!35}0.720 \\
     &0.95  & \cellcolor{red!50}\textbf{0.655} & \cellcolor{red!25}0.847 & \cellcolor{red!50}\textbf{0.283} & \cellcolor{red!10}0.653 \\
\end{tabular}
\caption{\textbf{Impact of the hyper-parameters on the performance.} This table shows results for the hyper-parameters: $k$ defining the grid level $\calG_k$, and $\tau_\text{corr}$, $\tau_\text{dens}$ and $\tau_\text{score}$, which control the filtering of correspondence, rotation hypothesis, and final 6D pose, respectively. Red values indicate the best parameters for precision-recall trade-off.}
\label{tab:ablation}
\end{table}

\textbf{Rotation grid levels ($k$).} Our method relies on an equi-volumetric partitioning of the $\mathrm{SO}(3)$ space with $\calG_k$. The volume of a bin of $\calG_k$ is directly related to its subdivision factor $k$.  We can notice that a too low $k$ reduces the recall, whereas a too high $k$ degrades the precision. This is because a low $k$ implies fewer bins which decreases the pose distribution coverage. At the opposite, when $k$ is high, the bins get smaller with fewer correspondences, decreasing the precision.

\textbf{2D-3D correspondences filtering ($\tau_{\text{corr}}$).} The 2D-3D matching process relies on the automatic computation of the descriptor similarity threshold, the latter depending on the parameter $\tau_{\text{corr}}$. We can observe from Table \ref{tab:ablation} that the influence of this parameter is limited, and the best precision-recall trade-off is reached with $\tau_{\text{corr}}=0.65$.

\textbf{Rotation hypotheses filtering  ($\tau_{\text{dens}}$).} The parameter $\tau_{\text{dens}}$ represents the minimal number of pose hypotheses per bin volume of  $\calG_4$. Smaller values retain more hypotheses, including from sparse regions, which increases recall but lowers precision. Larger values limit to hypotheses from dense regions, improving precision but lowering recall. We choose  $\tau_{\text{dens}}=10$ as a balanced  precision-recall trade-off. Still, the impact of this parameter is negligible.

\textbf{Pose scoring ($\tau_{score}$).} We examine the effects of filtering out outlier poses by score. We observe that lower thresholds retain more poses, increasing recall but lowering precision. While higher thresholds improve precision, they risk filtering out poses that are slightly misaligned, lowering recall. The threshold $\tau_{score}=0.9$ provides the best balance.

\subsection{Limitations}
Our evaluations are limited to the T-LESS dataset using the 6D pose distribution ground-truth introduced by BOP-Distrib \cite{bopd2024}, the only dataset currently available for this purpose. This limitation may introduce biases in the evaluations, highlighting the need for new datasets with accurate 6D pose distribution ground truth. Our method is trained on only 30 objects; a larger and more diverse dataset could allow generalization to unseen objects. During our evaluations, we observed that Corr2Distrib, as well as existing methods we compared to often over-spread the pose distributions when disambiguating elements become very small in the image; introducing a learning-based attention mechanism could address this. Additionally, we observe difficulties of the methods to handle the object texture inconsistency between the reference model and the images. Some objects in the real image display a texture that is missing in the CAD osbject model used to train the methods. Improved disentanglement of shape and texture could enhance robustness. Figure \ref{fig:limitations} illustrates these limitations.

\section{Conclusion}
\label{sec:conclusion}
In this paper, we present the first correspondence-based approach to estimate a pose distribution from RGB image and object model. Our approach takes advantages of descriptors ambiguity to recover all the valid poses that explain the observed image, especially in case of symmetries or occlusion-induced symmetries. Our evaluation, demonstrates that our solution outperforms state of the art for both 6D pose distribution and single pose estimation of a known object from an RGB image. These results underline the interest of considering pose distribution methods even if the final application only requires a single pose. We  believe this work demonstrates the potential of correspondence-based approaches for 6D pose distribution estimation, and we hope this paper  will inspire follow-up works.

This work benefited from the FactoryIA supercomputer financially supported by the Ile-de-France Regional Council. This project has received funding from the European Union’s Horizon Europe research and innovation program under grant agreement nº 101135708 (JARVIS Project).
\bibliographystyle{IEEEtran} 
\bibliography{main} 
\end{document}